\documentclass{article}

\usepackage{arxiv}
\usepackage[dvipsnames]{xcolor}
\usepackage[utf8]{inputenc} 
\usepackage[T1]{fontenc}    
\usepackage{hyperref}       
\usepackage{url}            
\usepackage{booktabs}       
\usepackage{amsfonts}       
\usepackage{nicefrac}       
\usepackage{microtype}      
\usepackage{lipsum}
\usepackage[pdftex]{graphicx}
\graphicspath{ {./images/} }
\usepackage{amsmath}
\usepackage{mathtools}
\usepackage{multirow}
\usepackage{multicol}
\usepackage{pifont}
\usepackage{xspace}
\usepackage[numbers]{natbib}
\newcommand{\cmark}{\ding{51}}
\newcommand{\xmark}{\ding{55}}

\usepackage{wrapfig}
\usepackage{authblk}

\def\L{\mathcal{L}}

\def\x{\mathbf{x}}
\def\y{\mathbf{y}}
\def\z{\mathbf{z}}
\def\vs{vs.\xspace}
\def\ie{i.e. \xspace}

\title{Discriminative Semantic Transitive Consistency for Cross-Modal Learning}

\author[1]{\textbf{Kranti Kumar Parida}}
\author[1,2]{\textbf{Gaurav Sharma}}

\affil[1]{Dept. of CSE, Indian Institute of Technology Kanpur, India}
\affil[2]{TensorTour Inc.\authorcr
	\{\tt kranti, grv\}@cse.iitk.ac.in}

\date{\vspace{-5ex}}

\begin{document}

\maketitle

\begin{abstract}
    Cross-modal retrieval is generally performed by projecting and aligning the data from two different modalities onto a shared representation space. This shared space often also acts as a bridge for translating the modalities. We address the problem of learning such representation space by proposing and exploiting the property of \textit{Discriminative Semantic Transitive Consistency}---ensuring that the data points are correctly classified even after being transferred to the other modality. Along with semantic transitive consistency, we also enforce the traditional distance minimizing constraint which makes the projections of the corresponding data points from both the modalities to come closer in the representation space. We analyze and compare the contribution of both the loss terms and their interaction, for the task. In addition, we incorporate semantic cycle-consistency for each of the modality. We empirically demonstrate better performance owing to the different components with clear ablation studies. We also provide qualitative results to support the proposals.
\end{abstract}


\section{Introduction}
With the rapid growth of digital devices and content, there is a plethora of data available today in many different modalities, e.g. audio, video, text, NIR, 3D and so on. The immediate challenge is to enable search and retrieval of appropriate content in all modalities given a query in any one modality. The methods for search and retrieval, when the query and the gallery both come from same modality, have been studied extensively and are now widely used in day to day life. However, the task of cross-modal retrieval, i.e. when the query and gallery are from different modalities, is more challenging and farther from widespread adoption. 

Cross modal retrieval involves a critical step of aligning different modalities in an intermediate space, after respective projections. It has been studied as an interesting task for a very long time and many different approaches have been proposed, e.g.\ canonical correlation analysis (CCA) \citep{thompson2005canonical} maximizes correlation between the modalities in the common space \citep{verma2017support}, auto encoders \citep{ngiam2011multimodal} align both modalities by enforcing reconstruction of data in one modality given the other modality as input.

Deep learning has been very successful in learning representations from raw input data, for different modalities alike, e.g.\ videos \citep{carreira2017quo}, text \citep{kim2014convolutional}, audio \citep{aytar2016soundnet}---all of these are processed with deep neural networks to give state of the art results in many uni-modal tasks. Approaches have also been proposed with deep learning for alignment of different modalities. The most common approach for the alignment is to use initial independent layers for each of the modalities followed by common layers \citep{aytar2017cross, zhen2019deep}. The independent layers act as non linear projections, for each modality into a common representation space, and the following common layers act as the shared classifier working with the common representation space for both modalities. The classifier learned in the common representation space with annotated data from both modalities, aligns the modalities to enable cross-modal retrieval in this space. To further strengthen the alignment, along with the classification loss, $\ell_2$ distance or negative cosine similarity minimizing losses, between the paired data in the different modalities are also used \citep{parida2020coordinated}. Such losses work with the projections in the common space and enforce that data points which correspond in the two modalities, e.g.\ audio and video from the same clip, are closer to each other either in absolute ($\ell_2$, cosine loss) or in relative terms (triplet loss)  wrt.\ the points which do not correspond, e.g.\ a video from a different clip.

We present a simple novel idea of discriminative semantic transitive consistency (DSTC), which is inspired by works on cyclic consistency \citep{zhu2017unpaired, dutta2019semantically} and is adapted for the task of cross modal (audio-visual or image-text) retrieval. We argue that the loss functions used in earlier works of enforcing the data points to lie close together in the common representation space might be too strict, given the final goal of semantic category-based retrieval. The case is similar for loss functions enforcing cyclic consistency, i.e.\ enforcing the representations to go to the exact same point when translated back to the originating modality. Instead, we propose to enforce a weaker form of correspondence using the individual representation space, i.e.\ we deem it to be sufficient if the projected points, from one modality, belong to the same class in the representation space of the other modality, as well as when translated back to representation space of the originating modality. In effect, the DSTC loss and its cyclic sibling, are satisfied if the audio and video data from the same clip do not necessarily coincide with each other, but do belong to the same class in both the representation space, and also maintain their class membership when translated back.

\begin{wrapfigure}{r}{0.5\textwidth} 
    \centering
    \includegraphics[width=0.5\textwidth]{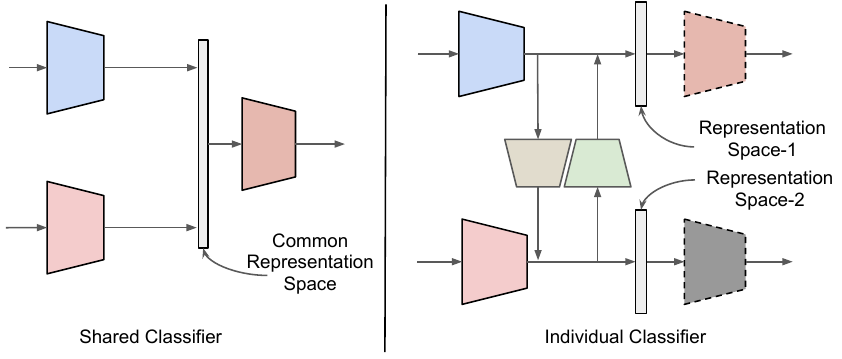}
    \vspace{-1 em}
    \caption{Block diagram showing the difference between existing (left) \cite{zhen2019deep,aytar2017cross} and proposed (right) approaches. In the existing approaches (left), there is a common representation space for both modality but in the proposed approach (right) each modality has individual representation space and the respective translators are used for aligning with the representation space of the other modality.
    }
    \label{fig:comp_block}
\end{wrapfigure}

The architecture we propose also differs from the popular existing architectures \citep{zhen2019deep,aytar2017cross}. While in the existing architectures, a shared classifier is learned in the common representation space, we learn individual classifiers for the modalities. Fig~\ref{fig:comp_block} compares the architecture we use to the traditional architectures. Instead of learning common representation space for both the modalities, we learn discriminative feature space individually for each of the modality first and then use translators for each of the modality to separately align one modality to the other. These translators create the bridges which enable cross-modal retrieval. They give the discriminative and transitive natures to the proposed semantic consistency, i.e.\ if an input $x:$`dog' in the audio modality, and $x:y$ via translation to the video modality, then $y:$`dog' by transitivity---the property preserved being that of semantic class discrimination. While the usual distance based losses act on the $x:y$ step. The DSTC acts on the $y:$`dog' step. Even within our framework, both the (cyclic) DSTC losses as well as pointwise correspondence based ones, after translation to respective modality spaces, can be used together. We investigate such combinations and show complementary strengths of both. We also give extensive empirical evaluations, quantitative and qualitative, to support our proposals. We will release code and trained models upon acceptance.

\section{Related Work}
\label{sec:related_work}

Our work is closely related to the topic of multimodal learning, cross-modal retrieval and the data and style transfer/translation.

\textbf{Multi-modal Learning.}
Multi-modal learning approaches can be broadly divided into two categories: (i) using already learnt model in one modality to learn or perform a task in other modality, (ii) using both the modalities to improve task performance cf.\ using a single modality.

In the first kind of approaches, many different tasks have been studied recently, such as learning audio representation from image \citep{aytar2016soundnet}, learning image representation from audio \citep{owens2016ambient}, recognizing emotions in audio by transferring knowledge from video \citep{albanie2018emotion}, pre-training action classification network for video by getting the labels from audio \citep{nagrani2020speech2action}, using video pre-trained network to track vehicles from audio \citep{gan2019self}. Cross-modal data is used in the framework of self-supervised learning in \cite{owens2018audio, arandjelovic2017look} as well to learn better representation in both the modalities by exploiting their correspondence in the data.

In the second kind, a variety of different approaches have been proposed, such as domain adaptation \citep{munro2019multi}, sound source separation \citep{gao2018learning, zhao2018sound}, depth estimation and visual navigation \citep{gao2020visualechoes, parida2021beyond}, binaural audio generation \citep{parida2022beyond}, zero-shot learning \citep{parida2020coordinated, mazumder2020avgzslnet} and person identification \citep{shon2019noise}, with the aim of improving performance using multiple modalities together. In another line of work cross-modal generation is performed \citep{hao2017cmcgan}, where the goal is to reconstruct one modality given other modality as input.

\textbf{Cross-modal Retrieval.}
Cross-modal retrieval approaches map data from both the modality onto a common representation to perform retrieval.  Such approaches can be broadly divided into three types on the basis of training strategy: (i) learning with full supervision \citep{jiang2017deep, zheng2016hetero, wang2017adversarial, zhen2019deep, wen2018disjoint, cao2017collective, luo2017simple, pereira2014cross}, (ii) zeros-shot retrieval (learning with limited supervision) \citep{parida2020coordinated, mazumder2020avgzslnet, dutta2019semantically, kumar2019generative, yelamarthi2018zero}, and (iii) self-supervised learning \citep{arandjelovic2017look, nagrani2018learnable, li2019coupled}. The proposed method is of the first kind, and requires full supervision for training for performing audio-to-video and image-to-text cross modal retrieval.

Closely related approaches for the problem, use shared classifier along with point wise correspondence  \citep{aytar2017cross, zhen2019deep} in the common representation space. Whereas, the proposed method uses modality specific classifiers with translator networks to bridge from one modality to the other for cross-modal retrieval. Our approach lends more flexibility to the projections from individual modality into the corresponding representation space cf.\ the shared classifier approaches. The proposed loss then enforces that even after translation to the other modality, the point retains its class membership, wrt.\ the classifier of the other modality. The main motivation here is that enforcing point wise correspondence is too strong condition for alignment, for the task of semantic cross-modal retrieval. For example, considering a `dog' barking audio/video sample---with pointwise correspondence the network will force the audio to be translated to that particular video, however it should suffice that the audio is translated to a video which belongs to the class `dog' as well. In one of the existing approach for cross modal retrieval using hash codes \citep{wu2018cycle}, the authors have used a similar approach of cycle consistency using pointwise correspondence to generate data from a common hash code. 

\textbf{Data Translation.} Image translation has been a popular topic recently, where a particular type of image is converted into a different type, e.g.\ day to night \citep{isola2017image}, gray to RGB \citep{zhang2017real} and summer to winter \citep{lee2018diverse}. Along similar lines, video-to-video synthesis \citep{wang2018video, chan2019everybody, zhou2019dance} approaches have also been proposed, where photorealistic videos are generated from a sequence of semantic segmented mask images. All these methods are trained in an adversarial fashion, however, most of them exploit the pointwise correspondence between the two type of data.

A variety of different approaches \citep{zhu2017unpaired, zhou2016learning, kulkarni2019canonical} are also used for translation of data points without any explicit correspondence annotated training data. These approaches uses the concept of cycle consistency to transfer the data from one type to another, i.e.\ the original input and the double translated output (from original space to the other space and back) should coincide with each other.
Cycle consistency has been used in many tasks such as image-to-image translation \citep{zhu2017unpaired}, canonical surface mapping \citep{kulkarni2019canonical}, cross-modal retrieval \citep{cornia2018towards}, zero-shot learning \citep{felix2018multi} etc.\ It enforces point-wise correspondence after double translation, i.e.\ to the other modality and back. Whereas, here we enforce semantic class consistency after completing the cycle, i.e.\ the point need not coincide with originating point, it is sufficient if the class membership is preserved, after double translation.
\section{Approach}
\label{sec:approach}

\begin{figure*}
    \includegraphics[width=\textwidth]{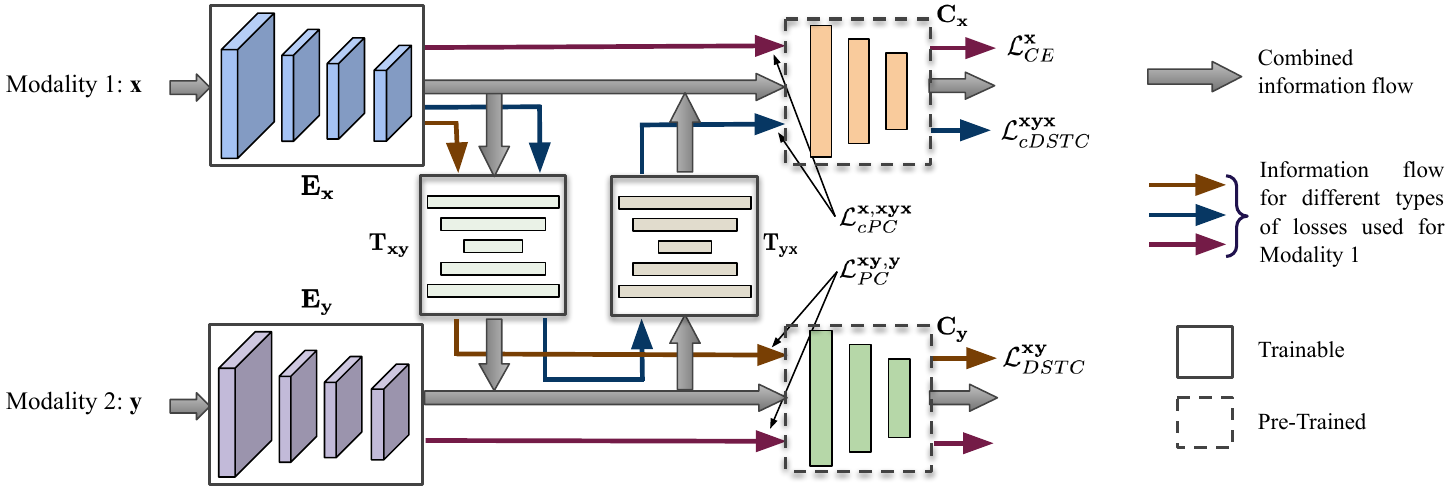}
    \vspace{-6mm}
    \caption{\textbf{Architecture, information flow, and losses.} The architecture consists, two each, of encoders $E$ which project the modalities onto a feature space, translators $T$ which can translate between modalities in this space, and classifiers which predict the classes for each modality. The cross modal retrieval is done in the representation space of the query modality using the translators, which is used to project the gallery examples onto the representation space of query modality. We show all the losses with their information flows, originating from modality 1. All of the losses shown (CE, DSTC, cDSTC, PT and cPT) have a symmetrical part originating from modality 2, and both parts are added to get the corresponding full losses as detailed in section below.}
    \label{fig:approach}
    \vspace{-1em}
\end{figure*}

\paragraph{Notations and Problem}
We work with paired data for both the modality, $\mathcal{D} = \{(\x_i,\y_i)\}_{i=1}^{N}$, $\x_i \in \mathbb{R}^{d_1}, \y_i \in \mathbb{R}^{d_2}$, $N$ being the total number of data points. Further, each pair of data has a class $\z_i = (z_{i1},\ldots,z_{iC}) \in \{0,1\}^C$ associated with it encoded as a one-hot vector, with $C$ being the total number of classes in the dataset. The problem of semantic cross modal retrieval is, given a query from one modality, retrieve data from the other modality. A retrieval result is valid when it has the same class label as the query, i.e.\ for the query $\x_i$, $\y_j$ is a valid retrieval if $\z_i = \z_j$. 

\begin{figure}
	\centering
	\includegraphics[width=0.8\columnwidth]{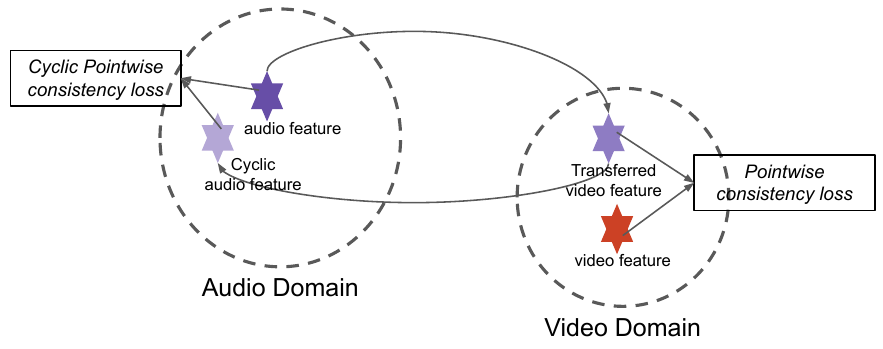}
	\caption{Illustration of transfer of features from one modality to other}
	\label{fig:loss_illustration}
\end{figure}

\subsection{Network Architecture and Losses}
\label{sec:losses}
The proposed network, Fig.~\ref{fig:approach}, contains six different sub-networks, two encoders $\mathbf{E_x}$ and $\mathbf{E_y}$ for encoding each of the individual modalities, two classification networks $\mathbf{C_x}$ and $\mathbf{C_y}$ for each of the modalities and two translation networks $\mathbf{T_{xy}}$ and $\mathbf{T_{yx}}$ for translating each of the modalities to the other respectively. Each of the sub-networks are MLPs themselves, the details of the number of layers and their sizes is given in the Experiments section. The network takes a pair of data points $\{(\x_i,\y_i)\}$ as the input while training and uses the following losses.

\paragraph{Cross entropy losses.} We use the standard cross-entropy losses for training the modality classifiers. In this loss, there are no interactions between the modalities, and the networks for individual modality are trained separately to improve the classification accuracy.
\begin{equation}
    \normalsize
\begin{split}
   \L_{CE}
   =-\frac{1}{N} \sum_{i=1}^{N} \sum_{c=1}^{C} z_{ic}\big[&\log  \left(\mathbf{C_x}(\mathbf{E_x}(\x_i))\right) +\\
   &\log \left(\mathbf{C_y}(\mathbf{E_y}(\y_i))\right) \big]
\end{split}
\end{equation}

\paragraph{Discriminative Semantic Transitive Consistency (DSTC) losses} The DSTC losses enforce that once an input from one modality is translated into the other modality, it maintains the same class, i.e.\ if $\x:$`dog' and $\x:\y$ by translation, then $\y:$`dog' as well, by transitivity of the property of discriminative class membership. Formally, the loss is given by
\begin{equation}
    \begin{split}
   \L_{DSTC}
   =-\frac{1}{N} \sum_{i=1}^{N} & \sum_{c=1}^{C} z_{ic}\big[\log \left(\mathbf{C_y}(\mathbf{T_{xy}}(\mathbf{E_x}(\x_i)))\right)\\ &+ 
   \log \left(\mathbf{C_x}(\mathbf{T_{yx}}(\mathbf{E_y}(\y_i)))\right) \big].
   \end{split}
\end{equation}

\paragraph{Cyclic DSTC (cDSTC) losses.} The cyclic versions of the DSTC loss ensures that when an input is double translated to the other modality and then back to the original modality, it maintains its class, i.e.\ if $\x:$`dog' and $\x:\y:\hat{\x}$ by cyclic translation, then $\hat{\x}:$`dog' as well.
\begin{equation}
    \small
    \begin{split}
   \L_{cDSTC}
   =-\frac{1}{N} &\sum_{i=1}^{N} \sum_{c=1}^{C} z_{ic}\big[\log \left(\mathbf{C_x}(\mathbf{T_{yx}}(\mathbf{T_{xy}}(\mathbf{E_x}(\x_i)))\right) \\ &+
   \log \left(\mathbf{C_y}(\mathbf{T_{xy}}(\mathbf{T_{yx}}(\mathbf{E_y}(\y_i)))\right) \big].
   \end{split}
\end{equation}

\paragraph{Pointwise consistency losses.}
Apart from the DSTC losses, we also use the paired data from both the modalities to enforce that the projection from one modality lies close to that from other, after translation, in the respective representation spaces, i.e.\

\begin{equation}
    \begin{split}
   \L_{PC}
   = \frac{1}{N} \sum_{i=1}^{N} &\bigg[ \lVert 
   \mathbf{E_x}(\x_i) - \mathbf{T_{yx}}(\mathbf{E_y}(\y_i)) \rVert_2^2 \\ &+
   \lVert 
   \mathbf{E_y}(\y_i) - \mathbf{T_{xy}}(\mathbf{E_x}(\x_i)) \rVert_2^2\bigg]
   \end{split}
   \label{eq:pc}
\end{equation}

\paragraph{Cyclic Pointwise consistency losses.} Similar to the cyclic DSTC loss, we enforce the pointwise consistency after double translation of a data point from one modality to the other and then back to the original modality, i.e.\ 
\begin{equation}
    \begin{split}
   \L_{cPC}
   = \frac{1}{N} &\sum_{i=1}^{N} \bigg[ \lVert 
   \mathbf{E_x}(\x_i) - \mathbf{T_{yx}}(\mathbf{T_{xy}}(\mathbf{E_x}(\x_i)) \rVert_2^2 \\ &+
   \lVert 
   \mathbf{E_y}(\y_i) - \mathbf{T_{xy}}(\mathbf{T_{yx}}(\mathbf{E_y}(\y_i)) \rVert_2^2\bigg]
   \end{split}
   \label{eq:cpc}
\end{equation}

In order to have an interpretable visual understanding of the pointwise loss functions, we give a schematic diagram in Fig.~\ref{fig:loss_illustration}. We also experiment with Cosine distance instead of Euclidean distance for both the losses $\L_{PC}$ and $\L_{cPC}$ in eq.~\ref{eq:pc} and eq.~\ref{eq:cpc} respectively. We do this by simply $\ell_2$ normalizing the vectors before the computing the Euclidean distance.

\subsection{Motivation of DSTC Loss}
\begin{figure}
    \centering
    \includegraphics[width=0.8\columnwidth]{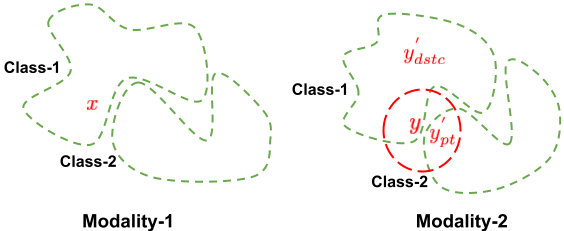}
    \caption{Illustration of discriminative vs.\ point wise loss. Let ${\color{red}x, y}$ be a paired data point from class-1 where ${\color{red}x}$ is from modality-1 and ${\color{red}y}$ from modality-2 respectively. ${\color{red}y^{'}_{dstc}}$ and ${\color{red}y^{'}_{pt}}$ represents the translation of ${\color{red}x}$ from modality-1 to modality-2 using DSTC and point-wise consistency respectively. The green dotted line ({\color{YellowGreen}- - -}) represents the class boundaries for different class in both modalities. The red dotted line ({\color{red}- - -}) represents the area where margin of point wise distance is minimum. It can be observed that the area within the green dotted line ({\color{YellowGreen}- - -}) are potentially correct region for translation using DSTC loss only. Similarly, the area within red dotted line ({\color{red}- - -}) are potentially correct regions for translation using pointwise loss only. Note here that using point wise loss only, doesn't respect the class boundaries which affects negatively for the task of semantic cross modal retrieval.} 
    \label{fig:illustration}
\end{figure}

We argue that point wise correspondence between data points, i.e.\ forcing paired data points from both modalities to be close to each other, from two different modalities is not optimal for the task of semantic cross-modal retrieval cf.\ DSTC. We provide a schematic diagram of this translation in Fig.\ref{fig:illustration}, where $\color{red}(x,y)$ represents the data points from first and second modalities respectively, and the green dotted line $({\color{YellowGreen}- - -})$ represents the boundaries of different classes in both modalities. In the figure, we point out that using discriminative loss only, gives translated data point ($y^{'}_{dstc}$) the flexibility to lie on the region of the same class in other modality. Similarly, using point wise loss only forces the translated data point ($y^{'}_{pt}$) to be within the small region around the original data in other modality (as mentioned with red dotted line in Fig.\ref{fig:illustration}). As using DSTC loss has potentially larger area for correct translation, it is a weaker form of translation. It can also be seen in Fig.\ref{fig:illustration} that enforcing only the pointwise loss only does not necessarily respect the class annotation in other modality.

\subsection{Information Flow and Training and Inference}
The final loss for training is given by a weighted average of the above losses, i.e.\
\begin{equation}
    \L = \L_{CE}+\alpha \L_{PC}+\beta \L_{DSTC}+\gamma \L_{cPC}+ \delta \L_{cDSTC}
\end{equation}
where $\alpha$, $\beta$, $\gamma$, $\delta$ are the hyperparameters used to control the relative weight of individual losses.

To train the network, we follow a 2-step approach, with the information flow for the different losses shown with different colors in Fig.~\ref{fig:approach}. In the first step, we individually train both the modalities for the task of classification by turning off the weights for the transfer module (i.e. using $\alpha$, $\beta$, $\gamma$, $\delta$ = $0$ ). 
In the second step, we learn to translate the modalities by jointly training the encoder and translator networks. The motivation of the architecture as well as the two step training procedure are closely connected to each other. While in previous works, e.g.\ \cite{aytar2017cross, zhen2019deep} the training forces both, alignment of two modalities in the common representation space as well as good classification with a shared classifier \emph{simultaneously}, we factorize it into two steps in the hope of making learning easier. Learning and freezing the classifiers in the first step, gives us a good individual representation space which is discriminative for the two modalities individually. In this step there is no alignment between the modalities and we achieve the alignment in the subsequent step by freezing the classifier and training the translators and encoders. We utilize the translator networks' capacity to do the alignment, such that the classification boundaries defined by the frozen classifier network are respected. The architecture and the training procedure allows us to separate the two aspects of learning alignment between the representations, and keeping them discriminative as well.

At test time, cross modal retrieval from one modality to the other is done using distance based scoring and sorting, for query $\x$ and gallery $\{\y_j\}$ as,
\begin{equation}
    \begin{split}
        s_j = \mathrm{score}&(\mathbf{E_x}(\x), \mathbf{T_{yx}}(\mathbf{E_y}(\y_j))),\\ 
        \textrm{output} &= \textrm{argsort}(\{s_j\})
    \end{split}
\end{equation}
where, scoring function can be $\mathrm{score}(a,b)=-\|a-b\|^2$ or $\cos(a,b)$.

\section{Experiments}
\label{sec:exp}

\begin{table*}[!htbp]
    \centering
    
    \resizebox{\columnwidth}{!}{
    \begin{tabular}{c|ccccc|ccc|ccc|ccc|ccc|ccc}
    \hline
    \textbf{Sl.}&\multicolumn{5}{c|}{\textbf{Losses}} &\multicolumn{3}{c|}{\textbf{Cos., Cos.}} &\multicolumn{3}{c|}{\textbf{Cos., Euc.}} &\multicolumn{3}{c|}{\textbf{Euc., Euc.}} &\multicolumn{3}{c|}{\textbf{Euc., Cos.}}& \multicolumn{3}{c}{\textbf{Class Average (Cos. dist)}} \\
    \textbf{No.}&\textbf{CE} & \textbf{PT} & \textbf{DSTC.} & \textbf{cPT} & \textbf{cDSTC} & {\textbf{A2V}} & {\textbf{V2A}} & {\textbf{Both}} & {\textbf{A2V}} & {\textbf{V2A}} & {\textbf{Both}}&{\textbf{A2V}} & {\textbf{V2A}} & {\textbf{Both}}&{\textbf{A2V}} & {\textbf{V2A}} & {\textbf{Both}}&{\textbf{A2V}} & {\textbf{V2A}} & {\textbf{Both}} \\
    \hline \hline
    1&\xmark&\cmark&\xmark&\xmark&\xmark&27.92&27.57&27.75&20.29&23.90&22.09&30.24&32.81&31.53&32.07&34.38&33.23&25.14&25.73&25.43\\
	2&\xmark&\xmark&\cmark&\xmark&\xmark&50.13&51.84&50.98&29.82&46.79&38.31&28.67&47.39&38.03&49.65&51.67&50.66&35.43&37.56&36.49\\
	3&\cmark&\cmark&\xmark&\xmark&\xmark&27.22&26.15&26.68&23.83&20.37&22.10&49.30&47.87&48.59&50.65&49.33&49.99&41.00&41.21&41.10\\
	4&\cmark&\xmark&\cmark&\xmark&\xmark&51.71&52.07&51.89&49.38&43.61&46.49&48.71&44.40&46.55&51.50&52.12&51.81&36.17&36.98&36.57\\
	5&\cmark&\cmark&\cmark&\xmark&\xmark&51.73&51.97&51.85&50.07&44.11&47.09&54.59&50.04&52.31&55.30&54.12&54.71&43.51&42.53&43.02\\
	6&\cmark&\cmark&\xmark&\cmark&\xmark&28.21&26.85&27.53&25.73&24.11&24.92&48.23&45.50&46.86&49.36&46.96&48.16&39.77&39.83&39.80\\
	7&\cmark&\xmark&\cmark&\xmark&\cmark&52.93&51.38&52.16&50.23&42.33&46.28&49.35&43.38&46.36&52.56&51.61&52.09&37.62&36.31&36.96\\
	8&\cmark&\cmark&\cmark&\cmark&\xmark&51.66&51.90&51.78&50.74&44.82&47.78&54.15&50.15&52.15&55.33&53.86&54.59&43.27&42.63&42.95\\
	9&\cmark&\cmark&\cmark&\xmark&\cmark&53.13&51.31&52.22&50.70&42.97&46.83&55.10&50.43&52.76&56.72&54.30&55.51&44.55&42.68&43.61\\
	10&\cmark&\cmark&\cmark&\cmark&\cmark&53.28&51.27&52.27&51.10&43.67&47.38&55.48&51.50&53.49&56.88&54.75&55.82&44.33&43.03&43.68\\
		\hline \hline
    \end{tabular}
    }
    \caption{Contribution of different loss terms on retrieval performance (mAP) for `val' set of \texttt{AudioSetZSL} using various distance methods at training and testing time. E.g.\ heading Euclidean, Cosine means that Euclidean distance was used during training and Cosine distance was used during testing.}
    \label{tab:av_zsl_dist_method}
\end{table*}

In our experiments we use two kinds of cross-modal dataset. The first kind of dataset contains audio and video modalities whereas the second kind contains image and text modalities. For audio-video, we use one of the recently proposed dataset, namely AudiosetZSL \citep{parida2020coordinated} for the task of multi-modal zero-shot learning involving both the audio and video modality. It is a multiclass extension of AudioSet dataset \citep{gemmeke2017audio} and is also large scale with around $130$k samples. We consider $23$ seen classes out of total $33$ classes available in the dataset as the unseen class is not available during the training or the pre-training of the network and this might affect the quality of the features and hence the performance of the network.

For image-text dataset, we use two most popular dataset, Wikipedia \citep{pereira2013role} and  Pascal Sentence \citep{rashtchian2010collecting} with $10$ and $20$ classes respectively.

\subsection{Datasets and Implementation Details}
\textbf{Audio-Video Dataset} AudiosetZSL \citep{parida2020coordinated} has  both audio and video modalities and is provided with train, val and test splits. We use the same same split for the `seen` classes' images. We use the features provided by the authors in \citep{parida2020coordinated}. The features for both audio and video are $1024$ dimensional each and are extracted using pre-trained networks. We also perform weighted random sampling for training as the dataset is highly imbalanced and follows a long-tailed distribution. We use 2 layer multilayer perceptrons (MLPs) for both encoders, single layer MLPs for classifiers, and symmetric hour glass type network for transfer modules with 3 hidden layers (see supplementary for details). 

We set all the losses to have equal weights, i.e. $\alpha, \beta, \gamma, \delta = 1.0$ by validation. We train the network with Adam optimizer and initial learning rate of $10^{-4}$ and subsequently changed to $10^{-10}$ after classifier training.

\textbf{Image-Text Datasets.}
We use \texttt{Wikipedia} \citep{pereira2013role} and \texttt{Pascal Sentence} \citep{rashtchian2010collecting} dataset. The former has $2866$ image-text pairs from $10$ classes while the latter has $1000$ pairs from $20$ classes. We extract the features (described in supplementary) following \cite{zhen2019deep}. We also fix the encoder and classifier architectures for both the modality following \cite{zhen2019deep} for a fair comparison. The encoders and classifiers are both single hidden layer MLPs, the translators are hour glass type networks with a single hidden layer as well.

We set the hyperparameters $\alpha=10^1$, $\beta=1.0$, $\gamma=10^3$, $\delta=10^2$ for \texttt{Wikipedia} and $\alpha=10^1$, $\beta=1.0$, $\gamma=10^{-2}$, $\delta=1.0$ for \texttt{Pascal Sentence} dataset. We use the learning rate of $10^{-4}$ for both the dataset. 

We set the hyperparameters using the performance on the val set. We also observe that the performance does not vary significantly with the change of values, i.e.\ sensitivity of the parameters are relatively low. We provide the performance of the method on the \texttt{Sentence Pascal} dataset for wide range of hyperparameters in the supplementary material to demonstrate the low sensitivity.

\subsection{Ablation Experiments}

We show the contribution of individual losses in the training of the network for the task of cross-modal retrieval for \texttt{AudiosetZSL} in Tab.~\ref{tab:av_zsl_dist_method}. We report the mean average precision (mAP) score used to evaluate the retrieval performance using two distance functions, Euclidean and Cosine 
at train and test time. Each column in Tab.~\ref{tab:av_zsl_dist_method} refers to one of the combinations used for distance calculation at train and test time respectively, e.g.\ Euclidean, Cosine means that Euclidean distance was used during training and Cosine distance was used during evaluation/testing. 

Since the \texttt{AudiosetZSL} dataset is highly imbalanced, we also report the class averaged mAP (AP is averaged for each query in the class to get mAP per class which is then averaged over all classes to get the class averaged mAP).

We observe that the retrieval performance is better when using Cosine distance at test time even if the training was done using Euclidean distance (eq.~\ref{eq:pc}, eq.~\ref{eq:cpc}). Similar observations have been reported in earlier works \cite{dutta2019semantically, yelamarthi2018zero} but potential explanations are missing. We analyze this behaviour at the end of this section.

\textbf{DSTC \vs PT} (rows 1 and 2): We observe that the DSTC loss consistently outperforms the PT loss in all the five metric ($27.75$ \vs $50.98$, $22.09$ \vs $38.31$, $31.53$ \vs $38.03$, $33.23$ \vs $50.66$, $25.43$ \vs $36.49$ row 1 and 2). This shows that the discriminative loss is more suitable than the pointwise loss for this task and also this observation is intuitive as there is no semantic information in case of pointwise loss but the discriminative loss enforces the semantic relationship between both the modality. 

\textbf{DSTC \vs PT with CE loss} (rows 3, 4 and 5): We now add CE loss individually to PT (row 3), DSTC (row 4) respectively, and also combine all three together (row 5). We observe that CE+DSTC (row 4) consistently outperforms CE+PT (row 3) while using cosine distance either in training or testing (col.\ 1, col.\ 2, col.\ 4) ($51.89$ \vs $26.68$, $46.49$ \vs $22.10$, $51.81$ \vs $49.99$) except when using euclidean distance both for training and testing (col.\ 3) ($46.55$ \vs $48.59$). This discrepancy is similar to that of Euclidean/Cosine difference mentioned earlier and discussed at the end of this section. We further observe that CE+DSTC (row 4) performs better than CE+PT (row 3) in individual cosine distance mAP ($51.81$ \vs $49.99$) but not in class average mAP ($36.57$ \vs $41.10$). The observed performance can be attributed to the fact that some examples are affected relatively more by the pointwise loss whereas some other examples are affected more by the discriminative loss. This observation is further reinforced by the fact that adding all the three losses (row 5) improves the performance significantly in almost all the cases ($51.85$ \vs $51.89$, $46.49$ \vs $47.09$, $46.55$ \vs $52.31$, $51.81$ \vs $54.71$, $36.57$ \vs $43.02$). The difference in performance from the global average to class average case for losses in rows 3 and 4 can be explained: as the dataset is highly imbalanced, possibly some larger class data are more dominated by the pointwise loss as compared to the discriminative loss.

\textbf{CE + PT + cPT \vs CE + DSTC + cDSTC} (rows 6 and 7): 
The performance of pointwise loss with the disrciminative loss along with the cycle terms shows similar trend as the similar loss combinations without the cycle losses (rows 3 and 4); the discriminative loss performs better in the individual case where as the pointwise loss performs better in the class average case.

\textbf{CE + PT + DSTC + cPT \vs CE + PT + DSTC + cDSTC} (rows 8, 9 and 10): We now show the impact of the two cycle loss terms on the overall performance. We observe that the addition of cPT decreases the performance or is at par with the baseline of previous three losses (row 8 \vs 5) for all the three distance metric ($51.78$ \vs $51.85$, $47.78$ \vs $47.09$, $52.15$ \vs $52.31$, $54.59$ \vs $54.71$, $42.95$ \vs $43.02$). The decreases in performance can be explained by the fact that the pointwise loss becomes too strict in matching the data point by point, i.e.\ it tries to match a particular 'dog' barking sound back to exactly the sound when double translated to video and back, as explained in related work section. We observe finally that adding all the losses improves the performance only marginally in all the case and class average distance ($43.68$ \vs $43.61$). This marginal improvement shows that the cyclic pointwise loss does not have much impact on the performance of the system. 

Since the method with Euclidean distance for training and Cosine distance for evaluation outperforms all other methods, we report all the following results with this setting.

\subsection{Euclidean \vs Cosine Loss}
\begin{figure}[h]
\centering
	\includegraphics[width=0.35\columnwidth]{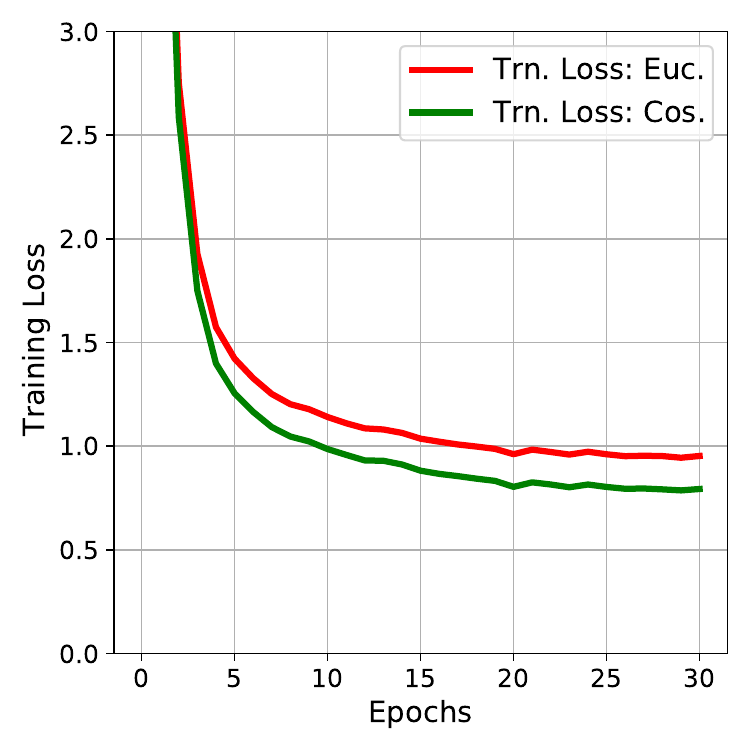}
	\includegraphics[width=0.35\columnwidth]{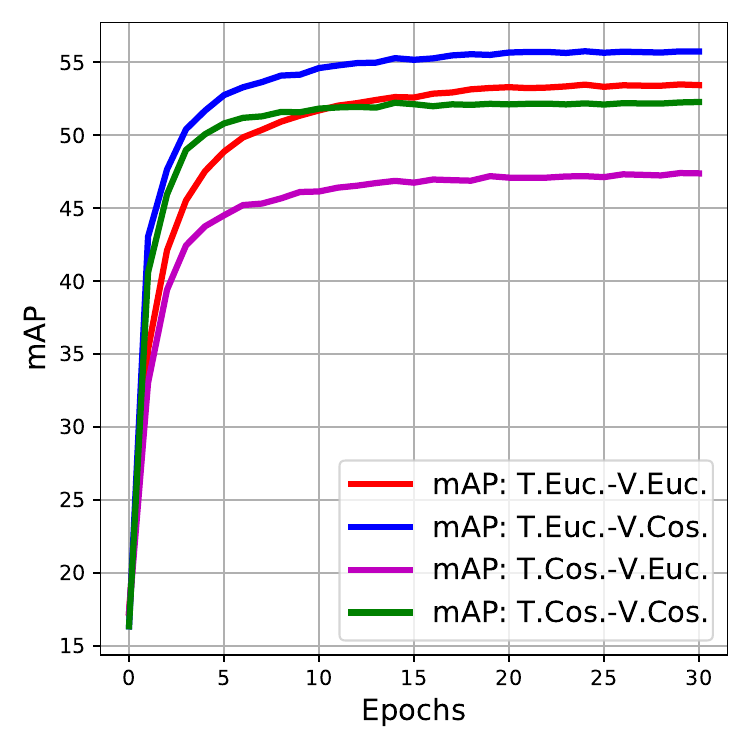}
	\vspace{-4 mm}
\caption{(left) Training loss  (right) Retrieval mAP.
}
\label{fig:euc_cos}
\end{figure}
Similar to \cite{dutta2019semantically, yelamarthi2018zero}, we observe that Cosine distance performs better even if the training was done using Euclidean distance. We observe from Fig.~\ref{fig:euc_cos} that the training loss using Cosine distance is lower cf.\ Euclidean distance. But the validation mAP using Cosine distance is better than using Euclidean with the same model, irrespective of the training distance used. This indicates Cosine loss in inherently better, but that training with Cosine loss overfits easily and degrades the performance. In practice using a model trained with Euclidean distance with Cosine distance during testing achieves a favorable balance.The cosine loss is equivalent to normalized euclidean distance with a scaling factor. We suspect normalization leads to better regularization and in turns provides a better generalization. This has only been empirically observed in earlier works \citep{dutta2019semantically, yelamarthi2018zero} as well and there is no theoretical insights for the same.

\begin{wrapfigure}{c}{0.6\linewidth}
	\centering
	\renewcommand{\arraystretch}{1.0}
	\begin{tabular}{c|ccc}
		\hline
		{\textbf{Method}}& {\textbf{Aud2Vid}} & {\textbf{Vid2Aud}} & {\textbf{Both}}\\
		\hline \hline
		pre-trained \cite{parida2020coordinated} & 3.61 & 4.22 & 3.91 \\
		GCCA \cite{parida2020coordinated, kettenring1971canonical} & 22.12 & 26.68 &24.4 \\
		CCA \cite{hotelling1992relations} & 33.55 & 32.60 & 33.07\\
		CJME \cite{parida2020coordinated} & 26.87 & 29.83 &27.95 \\
		AVGZSLNet \cite{mazumder2020avgzslnet} & 26.63 & 29.56 & 28.10\\
		MTFH(32) \cite{liu2019mtfh} & 54.48 & 52.52 & 53.50\\
		MTFH(64) \cite{liu2019mtfh} & 57.20 & 55.93 & 56.56\\
		DSCMR\textsuperscript{+} \cite{zhen2019deep} & 54.95 & 52.41 & 53.68 \\
		DSCMR\textsuperscript{+}( w/ class avg.)\cite{zhen2019deep} &40.21& 40.10& 40.15 \\
		\hline
		Ours & 57.81&55.09 &56.45 \\
		Ours(w/ class. avg.) & 41.21 &40.26 &40.73 \\
		\hline \hline
	\end{tabular}
	\caption{Retrieval performance (mAP) comparison of \texttt{AudiosetZSL} with existing methods}
	\label{tab:avzsl_prior_method}
\end{wrapfigure}

\subsection{Comparison with State-of-the-art Methods}
We now compare the proposed method with the existing state-of-the-art methods.

\textbf{Audio-Video Dataset:}
Tab.~\ref{tab:avzsl_prior_method} shows the results for \texttt{AudioSetZSL} dataset. We compare our approach to two baseline methods Canonical Correlation Analysis (CCA) and Generalized Canonical Correlation Analysis (GCCA). CCA learns a projection which maximizes the correlation of the two modalities in the common space. GCCA is the multi-set extenison of CCA where the correlation is maximized between all the sets. We report the numbers for these baselines from \cite{parida2020coordinated}. We also report the performance of two recently proposed zero-shot learning approaches \citep{parida2020coordinated, mazumder2020avgzslnet} that uses variant of triplet loss to align different modalities. All the results are reported on comparable experimental setup (i.e.\ on `seen' classes in the dataset).

We also show the result using one of the best performing text to image cross-modal retrieval method, DSCMR \citep{zhen2019deep}. As the original DSCMR has a different network structure, we modify it to match with that of ours which is tailored for the dataset (details in supplementary) to have a fair evaluation. We also compare our approach with a recently proposed hash code approach using matrix factorization, i.e. MTFH. In this approach binary hash codes are learnt for each modalities and then the same hash codes are used for retrieval. We show the results for two different length of hash codes, $32$ and $64$.

We observe that the proposed method with cosine distance for evaluation outperforms the pre-trained baseline, CCA, GCCA and the other methods by a convincing margin ($56.45$ vs $28.10$, $27.95$, $24.4$, $3.91$). We also observe that the proposed method outperforms DSCMR ($56.45$ vs $53.68$) which is the state of the art in text-image cross modal retrieval. Similarly, for the hashing based approach, MTFH, we observe that our approach outperforms it ($53.50$ vs.\ $5645$) for the hash code length of $32$. For hash code with length $64$, our approach almost similar to that of MTFH ($56.45$ vs.\ $56.56$). Finally, we also report the class average performance where the proposed method marginally outperforms DSCMR ($40.73$ \vs $40.15$).

\textbf{Image-text Datasets:} We report the comparison results for both \texttt{Pascal Sentence} and \texttt{Wikipedia} dataset with prior approaches, in Tab.~\ref{tab:pascal_prior_method} and Tab.~\ref{tab:wiki_prior_method} respectively. We report the mAP score for prior methods as provided by the authors in \cite{zhen2019deep}. Since there is no fixed split provided on the dataset, we perform the experiment with $10$ random train/test splits, and report the mean and standard deviation. We did the same for the state of the art DSCMR \citep{zhen2019deep} and MTFH \citep{liu2019mtfh} methods with the same random splits as well.

\begin{table}[!htbp]
	\begin{minipage}{.49\linewidth}
	\centering
	\resizebox{\columnwidth}{!}{
		\renewcommand{\arraystretch}{1.2}
    \begin{tabular}{c|ccc}
    \hline
    {\textbf{Method}}& {\textbf{Img2Txt}} & {\textbf{Txt2Img}} & {\textbf{Both}}\\
    \hline \hline
    CCA \cite{hotelling1992relations} & 22.5 & 22.7 & 22.6 \\
    JRL \cite{zhai2013learning}&  52.7 & 53.4 & 53.1 \\
    CMDN \cite{peng2016cross}& 54.4 & 52.6 & 53.5 \\
    CCL \cite{peng2017ccl}& 57.6 & 56.1 & 56.9 \\
    MvDA-VC \cite{kan2016multi} & 64.8 & 67.3 & 66.1 \\
    ACMR \cite{wang2017adversarial}& 67.1 & 67.6 & 67.3 \\
    DCCAE \cite{wang2015deep} & 68.0 & 67.1 & 67.5 \\
    DCCA \cite{andrew2013deep}& 67.8 & 67.7 & 67.8 \\
    DSCMR \cite{zhen2019deep} & 71.0 & 72.2 & 71.6 \\
    \hline
    MTFH(32)\textsuperscript{\textbf{+} \cite{liu2019mtfh}} & 58.07$\pm$2.93 & 62.85$\pm$1.00 & 60.46$\pm$1.75\\
    MTFH(64)\textsuperscript{\textbf{+} \cite{liu2019mtfh}} & 64.52$\pm$1.11 & 67.27$\pm$0.92 & 65.89$\pm$0.70\\
    DSCMR \textsuperscript{\textbf{+}} & 69.77$\pm$0.43  & 70.63$\pm$0.64 & 70.22$\pm$0.41 \\
    Ours \textsuperscript{\textbf{+}} &70.54$\pm$0.26 & 69.21$\pm$0.28 & 69.88$\pm$0.21\\
    \hline
    DSCMR  &60.82$\pm$3.19 &60.25$\pm$3.50 &60.54$\pm$3.09 \\
    Ours  &60.12$\pm$2.90 &60.62$\pm$2.99 &60.87$\pm$2.90  \\
    \hline \hline
    \end{tabular}
    }
    \vspace{-0.1em}
    \caption{Comparison of retrieval performance (mAP) for \texttt{Pascal Sentence} Dataset with existing methods. {\textbf{+}} denotes the method using features provided by the authors of \cite{zhen2019deep}.}
    \label{tab:pascal_prior_method}
\end{minipage}%
\hfill
\begin{minipage}{.49\linewidth}
\vspace{-2.1 em}
\resizebox{\columnwidth}{!}{
	\renewcommand{\arraystretch}{1.2}
    \begin{tabular}{c|ccc}
    \hline
    {\textbf{Method}}& {\textbf{Img2Txt}} & {\textbf{Txt2Img}} & {\textbf{Both}}\\
    \hline \hline
    CCA \cite{hotelling1992relations} & 13.4 & 13.3 & 13.4 \\
    MCCA \cite{rupnik2010multi} &34.1 & 30.7 & 32.4\\
    MvDA \cite{kan2016multi} &33.7 & 30.8 & 32.3\\
    MvDA-VC \cite{kan2016multi} &38.8 & 35.8 & 37.3\\
    JRL \cite{zhai2013learning} &44.9 & 41.8 & 43.4\\
    CMDN \cite{peng2016cross} &48.7 & 42.7 & 45.7\\
    DCCA \cite{andrew2013deep}&44.4 & 39.6 & 42.0\\
    DCCAE \cite{wang2015deep}&43.5 & 38.5 & 41.0\\
    ACMR \cite{wang2017adversarial}&47.7 & 43.4 & 45.6\\
    CCL \cite{peng2017ccl} &50.4 & 45.7 & 48.1\\
    DSCMR \cite{zhen2019deep} &52.1 & 47.8 & 49.9\\
    \hline
    MTFH(32)\textsuperscript{\textbf{+} \cite{liu2019mtfh}} & 45.27$\pm$1.40 & 42.40$\pm$1.12 & 43.83$\pm$0.79\\
    MTFH(64)\textsuperscript{\textbf{+} \cite{liu2019mtfh}} & 45.72$\pm$0.88 & 43.88$\pm$1.86 & 44.80$\pm$1.22\\
    DSCMR &44.68$\pm$1.57 &45.30$\pm$1.38 &45.00$\pm$1.42 \\
    Ours &47.74$\pm$0.94 &44.41$\pm$1.05 &46.08$\pm$0.95 \\
    \hline \hline
    \end{tabular}
    }
    \vspace{-0.1em}
    \caption{Comparison of retrieval performance (mAP) for \texttt{Wikipedia} Dataset with existing methods. }
    \label{tab:wiki_prior_method}
\end{minipage} 
\end{table}

\begin{figure*}
    \centering
    \includegraphics[width=0.956\textwidth]{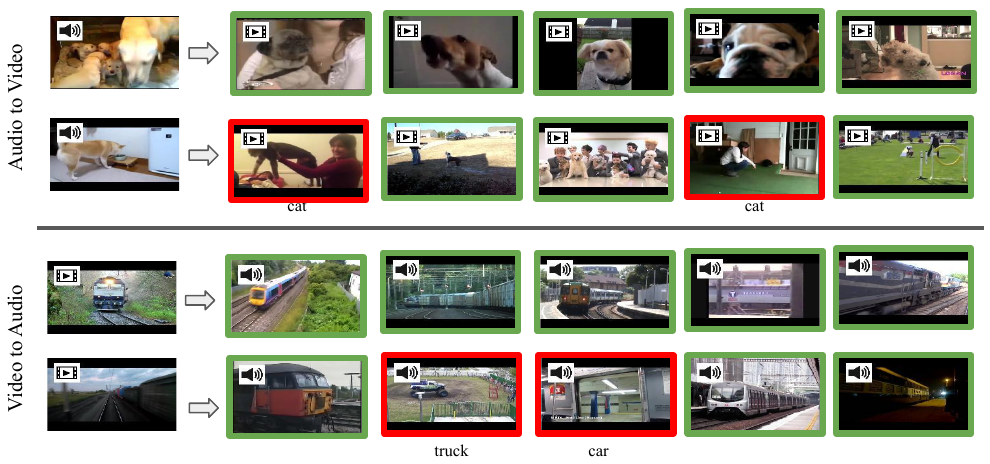} 
    \caption{Top-5 retrieval results for \texttt{AudioSetZSL} dataset. The first two rows are the results for audio to video retrieval and the next two are the results for video to audio retrieval, modality of the example is indicated by the icon in the top left of the image. The correct retrieval examples are marked by the green border where as the wrong ones with red. We note that the proposed method is able to perform retrieval with large amount of diversity in the data. See supplementary material for more results.}
    \label{fig:qual_res}
\end{figure*}

The authors of DSCMR \citep{zhen2019deep} have also provided the features for train and test split for \texttt{Pascal Sentence} dataset. We also report the mean and standard deviation of mAP score for the available features (marked with \textsuperscript{\textbf{+}}) in Tab.~\ref{tab:pascal_prior_method} for both the method (ours, DSCMR and MTFH). While we do not finetune our feature extraction networks on the target datasets, DSCMR \citep{zhen2019deep} seems to do that before training the main method. Hence we also compare with the finetuned features provided directly by them. We observe here that our approach performs marginally better than the best performing previous approach using extracted features both in the \texttt{Pascal Sentence} dataset ($60.87$ \vs $60.54$) and \texttt{Wikipedia} dataset ($46.08$ \vs $45.00$). We also observe that the proposed approach also significantly outperforms the hashing based approach for both the lengths of $32$ and $64$ for both \texttt{Wikipedia} ($43.83$ \vs $46.08$ and $44.80$ \vs $46.08$) and \texttt{Pascal Sentence} ($60.46$ \vs $69.88$ and $65.89$ \vs $69.88$).

The published numbers by other methods on Wikipedia dataset are higher, eg.\ DSCMR reports $49.9$ (while it obtains $45.00$ in our implementation). We believe that this is due to stronger features used by the previous approaches, which are unfortunately not publicly available for us to compare on. 

\subsection{Qualitative Results}
In Fig.~\ref{fig:qual_res}, we show some qualitative results for the \texttt{AudioSetZSL} dataset. We use a representative frame from the video to show the results for both audio and video. We observe that our model makes understandable mistakes in a few cases, e.g.\ in the second audio to video retrieval example, for the \texttt{dog} audio query a \texttt{cat} video is retrieved which looks similar in shape to that of a \texttt{dog}. In the video to audio retrieval, we find an interesting incorrect retrieval, the second query example of \texttt{train} video contains a retrieval audio example from the class \texttt{car} which is actually a train audio and is incorrectly labeled in the dataset. The same \texttt{train} video query also has an incorrect retrieval of \texttt{truck}, which is incorrect but is very similar (in audio modality). Due to space constraint, we provide the retrieval results for image to text and text to image in the supplementary material.
\section{Conclusion}
\label{sec:conclusion}
We proposed a novel framework for the task of cross-modal retrieval by aligning data from two different modalities. We proposed a Discriminative Semantic Transitive Consistency (DSTC) loss which ensures that the class label of the data remains the same even after transferring it to other modality, and after a second successive translation bringing it back to the original modality. The methods projects the modalities onto a representation space with individual modality classifiers, and has modality translator networks to enable cross-modal retrieval. We provided extensive ablation experiments to understand the contributions of the different components. We also compared quantitatively on three challenging public benchmarks with existing methods, and showed qualitatively that the method is capable of achieving diverse retrievals. We will release code and trained models upon acceptance.

\bibliographystyle{unsrt}
\bibliography{egbib}

\clearpage
\appendix
\section{Datasets and Features}
\label{sec:dataset}

In our experiments we used $2$ different kinds of cross-modal datasets. The first kind contains audio and video modality where as the second kind contains image and text modality.

\textbf{Audio-video Dataset:}
We use one of the recently proposed dataset, namely \texttt{AudiosetZSL} \cite{parida2020coordinated} involving both the audio and video modality for the task of multi-modal zero-shot learning. The features provided by the authors \cite{parida2020coordinated}, were extracted using a neural network with the I3D architecture for the videos pre-trained on the kinetics dataset \cite{carreira2017quo}, and a recently proposed audio classification network \cite{kumar2018knowledge} for audio. The audio network is not pre-trained with an auxiliary large dataset and is directly trained on the train set of \texttt{AudiosetZSL}. We use the $23$ `seen` classes out of total $33$ classes available in the dataset, which was split into `seen` and `unseen` classes for the zero-shot task.
We use the same train, validation and test split, within the `seen` classes' images as proposed in \cite{parida2020coordinated} and also perform weighted random sampling for training as the dataset is highly imbalanced and follows a long-tailed distribution. The portion of the dataset used contains $79795$, $26587$, $26593$ audio-video pairs in the train, val and test split respectively.

\textbf{Image-Text Datasets:} 
We use two popular datasets, \texttt{Wikipedia} \cite{pereira2013role} and \texttt{Pascal Sentence} \cite{rashtchian2010collecting} involving image and text modalities. We obtain the image features from the fc7 layer of VGG19 \cite{simonyan2014very} and the text features from the Sentence CNN \cite{kim2014convolutional} following \cite{zhen2019deep}. We average the sentence features over all the sentences as there were multiple sentence for a single input image example. The extracted features for image and text are of $4096$ and $300$ dimensions respectively. 

The \texttt{Pascal Sentence} dataset contains $1000$ image-text pairs from $20$ different classes with $50$ examples per class. All the prior works using the dataset have randomly split the data into $800$, $100$ and $100$ (with equal number of data points from each class) for train, val and test set respectively following \cite{feng2014cross}. As there is no unique split for all the three sets, the numbers reported with different methods can vary depending upon the random split of the dataset. In order to have a fair comparison we perform random split $k(=10)$ times and report the mean and standard deviations over all the runs for the test set. Apart from this, we also used the features for a fixed train and test split provided by the authors of \cite{zhen2019deep} (See Sec.~4.3 in the main paper).

The \texttt{Wikipedia} dataset, has a total of $2866$ images from $10$ different classes of which $2173$, $693$ image-text pairs belong to train and test sets respectively. In this case also there is no validation data, all the prior works follow \cite{feng2014cross} to split the original test data further randomly into test and validation set consisting of $462$ and $231$ data points respectively. Similar to the previous dataset, as there is no fixed val set, we randomly split the original test set $k(=10)$ times into test and validation sets and report the mean and standard deviations over all the runs, for the obtained test set.

\section{Implementation Details}
\textbf{Audio-Video Network.} 
We use a two-layered network each, for both the encoder (audio, video) networks ($\mathbf{E_x}$,$\mathbf{E_y}$)
with input and output node of size $1024$ and $256$ respectively. The hidden unit sizes are fixed to be $256$ and $512$ for the audio and video network respectively.
We use a single layer neural network as the classifier for each modality ($\mathbf{C_x}$, $\mathbf{C_y}$)
with input of size $256$ and output of size $23$. 
We use a symmetric hour-glass type network for the transfer modules ($\mathbf{T_{xy}}$, $\mathbf{T_{yx}}$). We use the same network structure for transfer module of both the modality, i.e.\ a MLP with 3 layers.
having same input and output dimension of $256$, $3$ hidden units of sizes $128$, $64$ and $128$ respectively. We use Batchnorm and ReLU non-linearity after each hidden layer for all the modules of the network.

\textbf{Image-Text Network.}
Similar to the audio-video network, the encoders and classifiers are both single hidden layer MLPs, the translators are hour glass type networks with a single hidden layer as well.
We fix the encoder and classifier architectures for both the modality following \cite{zhen2019deep}. The encoder architectures ($\mathbf{E_x}$, $\mathbf{E_y}$) are single layered neural network with $2048$ hidden units and $1024$ output units. The classifiers ($\mathbf{C_x}$, $\mathbf{C_y}$) are single layered neural network with $1024$ input units and $10$ output units. We use an hour-glass network for both the transfer modules ($\mathbf{T_{xy}}$, $\mathbf{T_{yx}}$) with input and output units of size $1024$ and single hidden layer of size $512$, and use BatchNorm and ReLU after each hidden layer.

\section{Qualitative Results}
\label{sec:supple_qual_res}

\begin{figure*}[h]
    \centering
    \includegraphics[width=\textwidth]{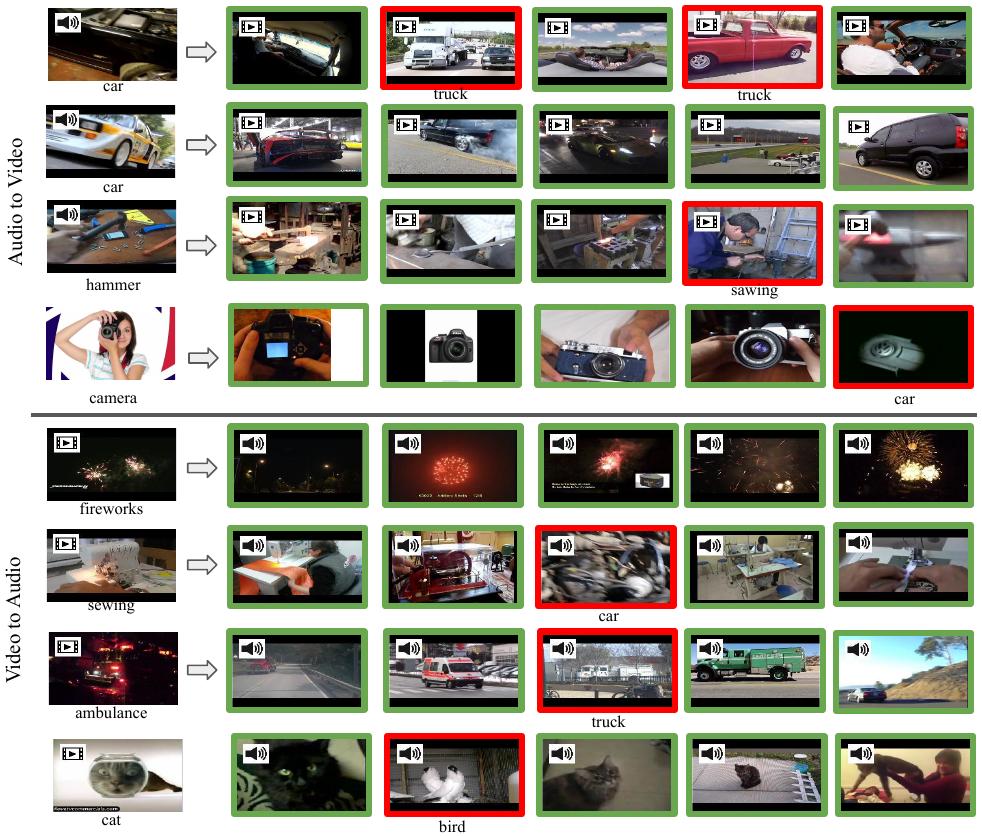} 
    \vspace{-8 mm}
    \caption{Top-5 retrieval results for \texttt{AudioSetZSL} dataset. The first three rows are the results for audio to video retrieval and the next three are the results for video to audio retrieval, modality of the example is indicated by the icon in the top left of the image. The correct retrieval examples are marked by the green border where as the wrong ones with red. We note that the proposed method is able to perform retrieval with large amount of diversity in the data.}
    \label{fig:qual_res_supple}
    \vspace{-2 em}
\end{figure*}

We provide here some additional qualitative results in Fig.~\ref{fig:qual_res_supple}. 

In addition, we request the readers to look at the videos available at
\url{https://krantiparida.github.io/projects/dstc.html} for a better understanding of the retrieval results. The discussion below refers to the videos available in the above link.

We provide four example video results each for audio-to-video and video-to-audio retrieval in separate files, as part of the supplementary material. The name of the files are self-explanatory. As our method performs cross-modal retrieval, we have switched off the other modality which was not used respectively in the query and the retrieved examples, \ie~audio is muted for all the retrieved examples in case of audio-to-video retrieval and for all the query examples in case of video-to-audio retrieval. Similarly, video modality is turned off for the other case, where for illustration we show a random frame from the video as a representative image for the entire duration (the selected frame appearance is not used for retrieval and is shown in the result just for illustration).

We mention some interesting observations from retrieval results below. \\

\textbf{Audio-to-Video Retrieval: }\\
\begin{itemize}
    \item[$-$] In \texttt{audio\_to\_video\_exp2.mp4}, the wrong retrieval results are from the class \texttt{truck} but looking at the video result, it can be seen that in one case there is actually a \texttt{car} along with the \texttt{truck} in the scene and in the other case the retrieval result is a `pickup truck` which is annotated as \texttt{truck} class in the dataset.
    
    \item[$-$] In \texttt{audio\_to\_video\_exp3.mp4}, the final retrieval result is from the class \texttt{car} instead of \texttt{camera}. But looking at the retrieval result we find out that the video contains only a \texttt{car} logo and it appears to be quite similar to that of a \texttt{camera} lens. 
   
    \item[$-$] Similarly in \texttt{audio\_to\_video\_exp4.mp4}, the wrong retrieval example is from the class \texttt{sawing} where as the query is from \texttt{hammer} class. Again looking at the results we find out that the act of performing \texttt{sawing} in the video is quite similar to that of \texttt{hammering}.
\end{itemize}

\textbf{Video-to-Audio Retrieval: }\\
\begin{itemize}
     \item[$-$] In \texttt{video\_to\_audio\_exp2.mp4}, the incorrect retrieval is from the class \texttt{truck} whereas the query is from \texttt{Ambulance}. But listening to the retrieval result we find out that it contains the ambulance siren as well---the retrieval result is semantically correct and the annotation is incorrect for that example.
     
     \item[$-$] Similarly for \texttt{video\_to\_audio\_exp3.mp4}, the query video is from the class \texttt{sewing} and the incorrect retrieval is from \texttt{car}. Again listening to the audio we find that it contains the sound of car engine, which in this particular case sounds similar to a \texttt{sewing} machine.
     
     \item[$-$] In \texttt{video\_to\_audio\_exp4.mp4}, all the retrieval results are correct and listening to them we find that they contain variety of train sounds, i.e.\ sound of horn blowing, sound of speeding train etc. This demonstrates that the method is able to generalize well enough to capture the variation in data within a class.
\end{itemize}

With these qualitative results we conclude that the method makes semantically sensible mistakes which are either due to the examples being very similar in the modality in question, or in rare case they have mistakes in annotations.

We provide here some qualitative results for Image-to-Text and Text-to-Image retrieval for \texttt{Sentence Pascal} dataset.

\begin{figure*}
    \centering
    \includegraphics[width=\textwidth]{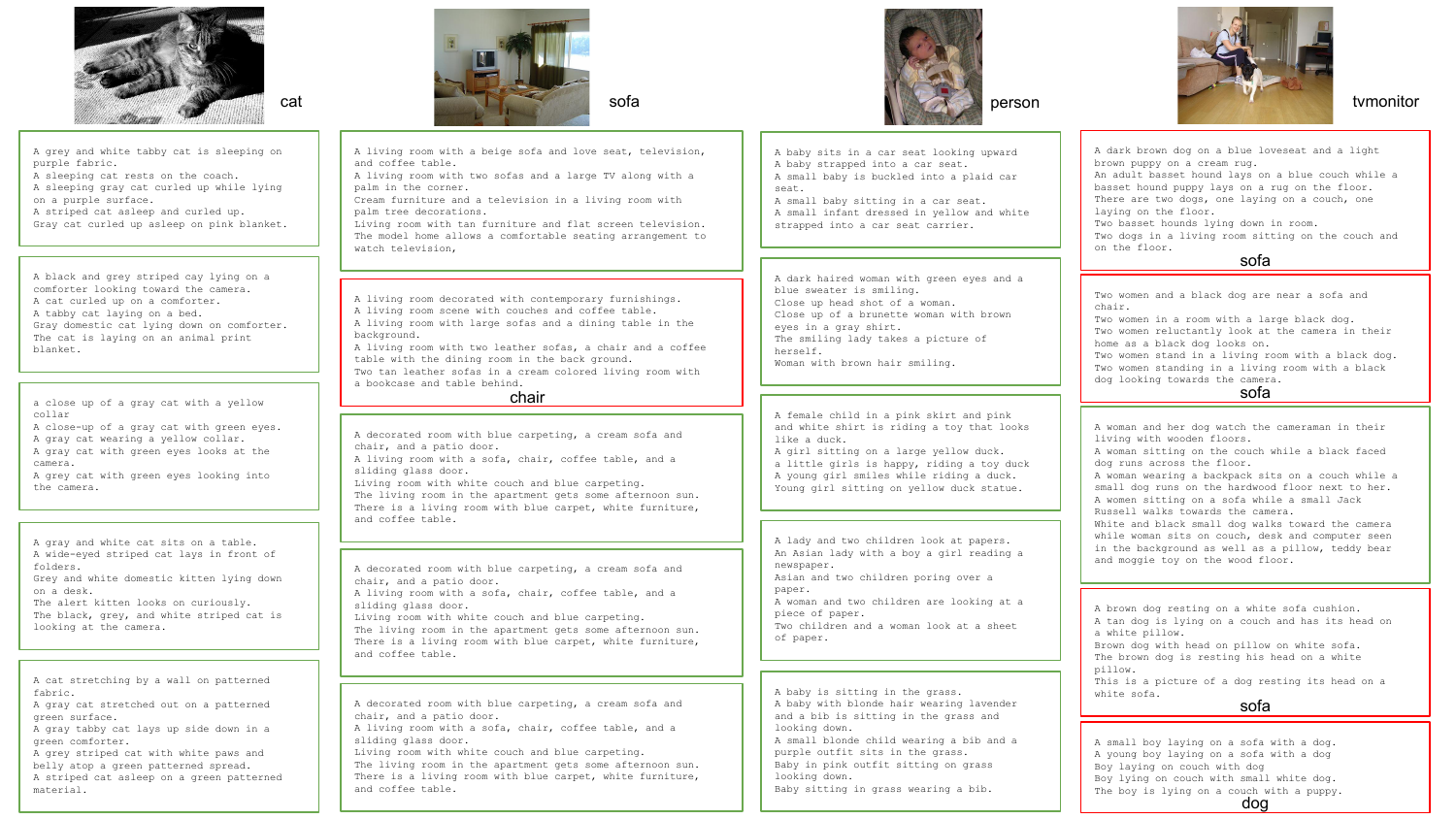}
    \caption{{Image to Text Retrieval.} Top 5 retrieval results for image to text retrieval for \texttt{Pascal Sentence} dataset. The top row represents the query image and the text in the five rows below represent the retrieval results. The correct retrieval results are marked with green border where as the incorrect ones are marked with red border. We observe that the results are semantically correct for most of the cases. For the incorrect ones, although the retrieval is not semantically correct, but the retrieval results are very similar to correct ones from other class.}
    \label{fig:img2txt}
\end{figure*}

\textbf{Image-to-Text Retrieval: }\\
We provide the results for image-to-text retrieval in Fig.~\ref{fig:img2txt}. The semantically correct retrieval are marked with green borders where as the incorrect ones are marked with red. From the results we observe that for most of the cases it produces semantically correct retrieval results and. For the incorrect ones although the retrieval results are not semantically correct but they individually have a high similarity with the query. For example, for the last retrieval results, the query is labelled as \texttt{tvmonitor} but the image contains sofa and dog as well. So the retrieval for this case contains text from those classes. Although the retrieval result can be considered as incorrect but this is a sensible mistake.   

\begin{figure*}
    \centering
    \includegraphics[width=\textwidth]{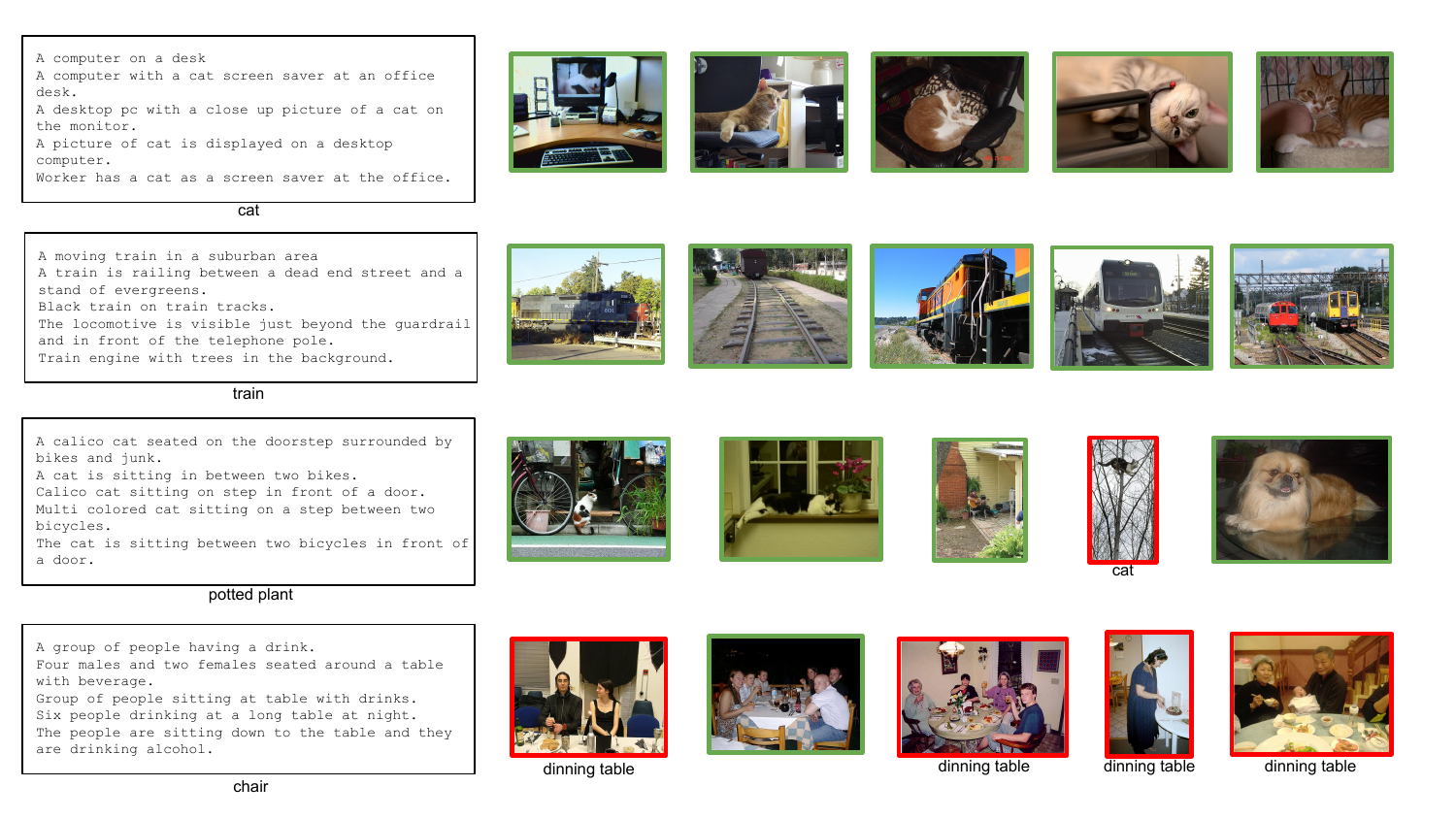}
    \caption{\textbf{Text to Image Retrieval.}Top 5 retrieval results for text to image retrieval for \texttt{Pascal Sentence} dataset. The left column represents the query text and the next five columns shows the retrieved image in decreasing order of similarity with the query. The correct retrieval results are marked with green border where as the incorrect ones are marked with red border. Similarly to that of image to text retrieval, we observe that the results our approach produces semantically correct results for most of the cases and also produces sensible failure cases.}
    \label{fig:txt2img}
\end{figure*}

\textbf{Text-to-Image Retrieval: }\\
We also provide some qualitative results for the case of Text-to-Image retrieval in Fig.~\ref{fig:txt2img}. Similar to Text-to-Image we mark the correct retrieval results with green boundary and the incorrect one with red. In this case also the failure cases retrieve meaningful results that are aligned with the query even though the class labels are different. In the last retrieval example, the query is from the class \texttt{chair} where as the retrieval results are mostly from the class \texttt{diningtable}, which is very similar to the class of \texttt{chair}.

\end{document}